\documentclass[sigconf]{acmart}

\usepackage{booktabs} 
\usepackage{tikz}
\usepackage{pgfplots}
\usepackage{amsmath}
\usepackage{amssymb}
\usepackage{graphicx}
\usetikzlibrary{matrix,calc,shapes,chains,positioning,decorations.pathreplacing,arrows}

\setcopyright{none}

\acmDOI{}

\acmISBN{}

\acmConference[arXiv Preprint]{}{Dec 2018}{}
\acmYear{2018}

\acmArticle{}
\acmPrice{}

\begin{document}
\title[Spartan Networks]{Spartan Networks: Self-Feature-Squeezing Neural Networks for increased robustness in adversarial settings}
\subtitle{}

\author{Fran\c{c}ois Menet}
\authornote{Corresponding author}
\affiliation{%
  \institution{Polytechnique Montr\'eal}
  \city{Montr\'eal}
  \state{QC}
}
\email{francois.menet@polymtl.ca}

\author{Paul Berthier}
\affiliation{%
  \institution{Polytechnique Montr\'eal}
  \city{Montr\'eal}
  \state{QC}
}
\email{paul-2.berthier@polymtl.ca}

\author{Michel Gagnon}
\affiliation{%
  \institution{Polytechnique Montr\'eal}
  \city{Montr\'eal}
  \state{QC}
}
\email{michel.gagnon@polymtl.ca}

\author{Jos\'e M. Fernandez}
\affiliation{%
  \institution{Polytechnique Montr\'eal}
  \city{Montr\'eal}
  \state{QC}
}
\email{jose.fernandez@polymtl.ca}

\renewcommand{\shortauthors}{F. Menet et al.}

\begin{abstract}
Deep learning models are vulnerable to adversarial examples which are input samples modified in order to maximize the error on the system. We introduce Spartan Networks, resistant deep neural networks that do not require input preprocessing nor adversarial training. These networks have an adversarial layer designed to discard some information of the network, thus forcing the system to focus on relevant input. This is done using a new activation function to discard data. The added layer trains the neural network to filter-out usually-irrelevant parts of its input. Our performance evaluation shows that Spartan Networks have a slightly lower precision but report a higher robustness under attack when compared to unprotected models. Results of this study of Adversarial AI as a new attack vector are based on tests conducted on the MNIST dataset.
\end{abstract}

%
%



\maketitle

\section{On the motivations for Robust Deep Learning}
\subsubsection*{Neural networks and deep learning}
Neural networks are machine learning algorithms that are mainly used for supervised learning. They rely on stacked layers of neurons. These stacked layers take a fixed-length input tensor and generate another output fixed-length tensor. The input/output function is differentiable relatively to its weights. By using gradient-based optimisation on large datasets to update the weights ---a process also refered to as ``training"--- the various layers generate output tensors whose values are a pattern-matching-value of their input, allowing these algorithms to detect features. Deep Learning consists of stacking a large amount of layers, allowing a neural network to extract features of features and thus grasp more abstract and complex characteristics from the input. Deep Neural Networks (DNN) have drawn a lot of attention recently.\\

In the last decade, DNN have revolutionized the automation of perceptive tasks in various domains, especially in computer vision. Deep Learning is increasingly used in autonomous driving software \cite{bojarski_end_2016}, malware analysis \cite{xu_neural_2017} (with attacks already implemented on detection systems \cite{hu_black-box_2017,grosse_adversarial_2016}), and fake news detection \cite{noauthor_fake_nodate}.\\

These technological improvements are already being used in safety-critical environments such as vehicles or factories, where automated driving for the former and predictive maintenance for the latter may save millions of dollars and thousands of lives.\\

\subsubsection*{New Attack Vector}
In late 2013, Szegedy et al.\cite{szegedy_intriguing_2013} discovered a new kind of vulnerability in DNN: given a neural network, the authors propose an alogrithm to generate samples that are missclassified while retaining their \emph{meaning} to the human cognitive system. This groundbreaking discovery created an entirely new threat model against machine learning powered applications. Several attacks have then been discovered. As an example, The Fast Gradient Sign Method (FGSM) \cite{goodfellow_explaining_2014} gives a reliable attack method against DNN. This method exploits the neural network instability to small adversarial input variations. This means that two samples that slightly differ from each other can be classified differently, even if they are indistinguishable for a human observer. The FGSM iteratively adds or substracts a small value $\epsilon$ to each element of the input tensor. This simple method yields surprisingly powerful attacks: on FGSM-generated adversarial samples\cite{goodfellow_explaining_2014}, precision of these algorithms can drop from 99\% to less than 20\%. \\

While classfier-evasion techniques are widely studied, DNN create an original problem as its lack of explainability and high sensitivity make them vulnerable to undistinguishable evasive samples. The performance of DNN also works against them: users tend to trust these systems because of their comparable performance to humans on some narrow tasks. This new attack vector does not exploit vulnerable code logic, but abuses users' trust in the classifier's complex decision boundaries.\\

Potentially any deep learning model can be vulnerable to this kind of attack, which is hard to detect, prevent, and whose impact will only grow in the upcoming years.
\subsection{Adversarial Examples and the Clever Hans Effect}
The vulnerability of these models to adversarial inputs brings forth another issue. If deep learning algorithms are seeing patterns that a human being can not see \textit{under normal conditions}\cite{elsayed_adversarial_2018}, this means that the models suffer from the so-called \emph{Clever Hans effect}. This term, popularized by \citet{papernot2018cleverhans}, comes from a horse that was deemed capable of complex arithmetical operations, where in fact the animal was guessing the actual answer from the unconscious behaviour of the audience.

The Clever Hans effect is the product of three observations~:\begin{enumerate}
\item Samples that are not generated by adversaries (i.e.~\emph{sane samples}) are almost always well classified by deep learning algorithms.
\item Samples that are generated by adversaries (i.e.~\emph{adversarial samples}) are likely to be misclassified by deep learning algorithms.
\item Human observers can still classify samples generated by adversaries \textit{given enough time} \cite{elsayed_adversarial_2018}.
\end{enumerate}

We can thus deduce that: 
\begin{itemize}
\item deep learning algorithms focus on features that are not essential to the true class of a sample.
\item features captured by deep learning algorithms during training are not the same as those learned by human beings over the course of their lives.
\end{itemize} 

Although the work of \citep{elsayed_adversarial_2018} shows common vulnerabilites between learning algorithms and a biological learning process, we can still state that the current learning behaviour of these algorithms do not capture the actual \emph{meaning} carried by the sample.

\subsection{Overview of our Work}
In our work, we consider that the current training dynamics of DNN creates their sensitivity to adversarial input. We will hypothesize on the characteristics of the algorithm causing devious behaviour, and we will create a minimal deep learning algorithm constrained to function without them.\\

We will study adversarial image classification as this community has created a large number of attacks and defenses. This plethora of research is due to the simplicity of the image space topology. For example, if one changes a pixel, or slighly varies the colors of various pixels in a panda image, the image still resembles a panda. On the other hand, trying to preserve the semantics of a ASM-x86 code while randomly changing a few lines of code would yield very different results.\\

Thus, we define Spartan Networks, and apply the general framework to Convolutional Neural Networks (CNN), as they are the state-of-the-art for image classification.\\

Our hypotheses on the current CNN are as follows:
\begin{enumerate}
\item The behaviour of the function $F_{nn}$ approximated by the CNN is locally linear, thus allowing an attacker to easily explore the system's state space.
\item Around sane datapoints, the function $F_{nn}$ is sensitive to features that do not make sense to a human observer.
\end{enumerate}

As we consider perturbations to be unnecessary information taken into account by the network, we try to create a network that learns to ignore parts of the information it is given.

In this paper, we propose a new type of DNN, the Spartan Networks, based on two conflicting elements forced to collaborate during the training phase:
\begin{itemize}
\item Firstly, the \emph{filtering layers} severely reduce the amount of information they give to the next layer. They are constrained to output the lowest amount of information through a \emph{filtering loss}.
\item Secondly, the other layers connected to the previous ones constitute a standard CNN trying to rely on the information given to minimize its training loss.
\end{itemize}

These two parts are competing against each other: if the filtering layer destroys all the information, the filtering loss is low, but the network cannot train, and thus the training loss is high. On the opposite, a CNN training given unlimited information allows the network to train efficiently, reaching a low training loss, but increasing the filtering loss.\\

This construction thus constitutes a \emph{self-adversarial} neural network. The weighted sum of those two losses forces the filtering layer to find the vital pieces of information the rest of the network needs to successfully train. The network thus learns to focus on less information, and selects more relevant features in order to maximise its performance.

This paper is organized as follows. A taxonomy of attacks specific to machine learning-powered applications is given Section 2. We describe the various test-time adversarial attacks among the aforementioned attacks in section 3. We present the various defense attempts in the litterature in Section 4. We explain the motivations for Spartan Networks in Section 5. We define candidate implementations of our proof-of-concept (PoC) in Section 6. We evaluate and discuss the results in Section 7. In Section 8, we will balance out the performance drop with the robustness gain to evaluate the relevance of Spartan Networks. We will conclude and present future work in Section 9.


\section{Attacking a Neural Network}
\subsection{Threat Model}
There are two main attack vectors available to an adversary to hinder a DNN's performance: Train-Time (or Poisoning), and Test-Time Adversarial Attacks.
\subsubsection*{Train-Time Attack}
This attack aims at modifying a dataset, by either adding patterns into existing samples, or adding new samples. The attacker's intent is to manipulate a deep neural network's training on this dataset in order to:
\begin{itemize}
\item create a \emph{backdoor}, which is a range of samples that are missclassified by the target network when it has been trained on the poisoned dataset. As an example, this could allow an attacker to evade malware detection for a specific type of malware, meaning the system would catch other malware with high accuracy, increasing user's trust in the system, but let the attacker's malware through.
\item diminish the overall accuracy of the neural network when trained on the poisoned dataset. As an example, an attacker could feed poisonned data that would cause an algorithm to extract the wrong features, causing additional data curation cost.
\end{itemize} 
Plausible attack scenarios within this threat model are given by Gu et al. \cite{gu_badnets_2017}.
\subsubsection*{Test-Time Attack}
In this type of attack, the neural network is already trained, its parameters are frozen and the attacker can only move within the space of all possible inputs. The attacker's objective is to find a sample $x_{adv}$ that is:
\begin{itemize}
\item \emph{close} to a sample $x$ correctly classified in class $y$
\item classified in a different class $y_{adv}\neq y$
\end{itemize}

The main hypothesis here is that when two samples are \emph{close}, their meaning stays the same to a human observer.

If the attacker succeeds most of the time for adversarial samples reasonably \emph{close}, it can reliably output samples indistinguishable from sane samples, that a classifier would fail to categorize well. For example, an attacker trying to bypass an automated content filter could take shocking pictures, slightly modify them, and succesfully bypass the filter. As the samples are close to their original counterpart, their meaning would be preserved.

More formally, we replace the \emph{closeness} by a distance between input and adversarial input, and get a constrained optimization problem:

Given a classifier $f$, a distance $\mathcal{D}$, a perturbation amplitude budget $\epsilon$, an attacker tries to find the minimal $\lambda$ on a sample $x$, with a ground truth label $y$ such that :

\begin{equation}
\label{opti}
\begin{aligned}
&f(x) = y \\
&f(x+\lambda) = y_{adv} ,y_{adv}\neq y \\
&\mathcal{D}(x,x+\lambda) < \epsilon
\end{aligned}
\end{equation}

Note that without the $\epsilon$ budget constraint, we could generate misclassifications by using a correctly classified input from another class.\\

In most cases, this distance is replaced by a standard $\mathcal{L}_{n}, n \in \{0,1,2\}$ or $\mathcal{L}_{\infty}$ norm for the perturbation $\lambda$. We depict below the various standard norms, with $\lambda=(\lambda_{i})_{i \in [\![1,k]\!]}$
\begin{center}

\begin{tabular}{|c|c|}
\hline 
Norm name & Mathematical expression \\ 
\hline\hline 
$\mathcal{L}_{0}$ & $\mathcal{L}_{0}(\lambda)=\#\{i|\lambda_{i}\neq 0\}$ \\ 
\hline 
$\mathcal{L}_{1}$ & $\mathcal{L}_{1}(\lambda)= \sum_{i}\vert\lambda_{i}\vert$ \\ 
\hline
$\mathcal{L}_{2}$ & $\mathcal{L}_{2}(\lambda)= \sqrt{\sum_{i}\lambda_{i}^{2}}$ \\
\hline 
$\mathcal{L}_{n}$ & $\mathcal{L}_{n}(\lambda)= \sqrt[n]{\sum_{i}\lambda_{i}^{n}}$ \\ 
\hline 
$\mathcal{L}_{\infty}$ & $\mathcal{L}_{\infty}(\lambda)= \max_{i}(\vert\lambda_{i}\vert) $ \\ 
\hline 
\end{tabular} 

\end{center}

In the rest of this paper we will focus on the Test-Time Attack, as this is where the attack surface is the largest.

\section{Threat Models for Test-Time Attacks}
There are various ways to attack a deep learning algorithm at test time. We contextualise the attacks by defining the different attack scenarios.\\

In order to understand the implications of adversarial examples, we will give a security equivalent of our attack scenario through a simple example. This will link the new attack vector to otherwise known attack vectors in the domain of information security.
\subsection{Attack scenario}
In order to place ourselves in an information security context, we may consider a simple setup: 
\begin{itemize}
\item The defender runs a check digit identification software powered by DNN on the target computer. It is the only program available on this system.
\item The attacker can send samples of checks. They have access to a very small digit database. Their original check is correctly classified with the right amount, but they aim at maliciously changing the input image in order to get a different check value. 
\item Opportunity: Through two accounts, the attacker can send checks to himself. They can thus check if the system is vulnerable.
\end{itemize}

\subsection{Attacker's access to knowledge}

As with any attack, attacks on Deep Learning vary on the knowledge the attacker has on the system prior to the attack.

\paragraph{White-Box Attacks}
In White-Box mode, the attacker has access to every parameter of the Neural Network. In this scenario, the attacker has a \emph{read-only} access to the all parameters of the algorithm by the means of an ill-protected file.
\paragraph{Gray-Box Attacks}
In Gray-Box mode, the attacker has access to some parameters of the Neural Network, while some remain unkown to them. A common scenario is that the attacker knows the structure of the Deep Neural Network, but not the parameters. In this case, the attacker has managed to get a restricted account on the server, showing only the code used to train, but not the admin-owned weights parameter.
\paragraph{Black-Box Attacks}
In Black-Box mode, the attacker has no access to the Neural Network other than through its inferences, like any normal user. In our scenario, this means that the attacker has no access to the server other than the check submission point.

\subsection{Attacker's Objective}

Regardless of the operational objectives, we give the two main \emph{technical} objectives attackers can set when they attack deep learning algorithms.

\paragraph{Untargeted Attack}
When the attack is untargeted, the attacker's objective is to create \emph{any} misclassification they can, with no control over the $y_{adv}$ in equation \ref{opti}. In our scenario, the attacker tries to trigger a misclassification from the real digit class.

\paragraph{Targeted Attack}
When the attack is targeted, the attacker's objective is to trick the algorithm into classifying the input in a class \emph{chosen by the attacker}, with complete control over the $y_{adv}$. In our scenario, and for simplicity, the attacker tries to classify all digits as a 9 digit. We will consider that it can only attain this goal through a \emph{targeted attack}.

We summarize the attack characteristics with respect to our scenario below :

\begin{tabular}{|c|c|}
\hline 
Attack & Attacker in the Scenario \\ 
\hline \hline
White-Box & Already has access to privileged information \\ 
\hline 
Gray-Box & Already has access to restricted information \\ 
\hline 
Black-Box & Has no access to any information \\ 
\hline 
Targeted & Wants to get a 9999 \$ check \\ 
\hline 
Untargeted & Wants to modify the check's value \\ 
\hline 
\end{tabular} 
\vspace{5pt}

We create an array from the last two sections, and get the following example attackers. U is for Untargeted, T for Targeted :\\

\begin{tabular}{|c|c|c|c|}
\hline
&White-Box & Gray-Box & Black-Box \\ 
\hline\hline
U&Disruption & Rogue Employee & DoS extortion\\ 
\hline 
T&Smash-and-Grab & Fraud & Theft \\ 
\hline 
\end{tabular} 
\vspace{5pt}

We describe the scenarios:

\paragraph{Disruption}
When the attacker has access to every information in the neural network, we consider that they already have access to a privileged account. For the sake of this example, we consider the case of an attacker with root access that wants to minimize its footprint. By allowing themselves to only read the parameters, the attacker can create a copy of the neural network then use it to disrupt business by forcing bank employees to manually review all checks emitted by the attacker.

\paragraph{Rogue Employee} 
In this context, an IT employee with a restricted account and thus some information on the neural network could trigger a misclassification on the system, and disrupt business by creating subtle modifications on all the checks submitted through its interface.

\paragraph{Denial of Service (DoS) extortion}
When the attacker has no access to the system, any misclassification from the original class will allow an attacker to get some form of disruption. If an attacker prints check paper with adversarial patterns on it, they can extort money from the bank by threatening to cause user mistrust and employee time to manually review checks.

\paragraph{Smash-and-Grab}
When the attacker is given White-Box access through a visible root access, they can create a base of malicious digit samples allowing them to transform any 4-digit check into a 9999\$ check. From this, the attackers could submit some malicious checks and take their money whenever possible. 

\paragraph{Fraud}
In this case, the attacker could be another rogue employee trying to do more than disruption by targetting all the digits to be 9.

\paragraph{Theft}
This is the worst scenario. If successful, the attacker can create a target misclassification with only access to the computer, completely controlling the value of the check.\\

These attacks could be seen as less risky variants of physically tampering the check's value. If caught, the attacker can blame the system, by arguing that a human reviewer can perfectly see the check's actual value. As these attacks are almost riskless and costless for attacker but disruptive for the defender, these AI attacks constitute an interesting field of study from an information security perspective.

These attacks are not limited to the banking system. As an example, deep learning algorithms are currently used in autonomous driving \cite{bojarski_end_2016}, malware analysis (with already an arms race between attacks and defenses \cite{grosse_adversarial_2016,xu_neural_2017,hu_black-box_2017}), healthcare\cite{esteva2017dermatologist}, and fake news detection\cite{noauthor_fake_nodate}.

\subsection{Main attacks implemented against Deep Learning}
For each attack, we will give a formal mathematical definition. We will then write an equivalent attack to understand the inner workings of the algorithm.

\paragraph{First attack discovered}
A lot of attacks aiming at generating adversarial samples have been developped in the last five years, since the discovery by \citet{szegedy_intriguing_2013} of a L-BFGS-driven line-search attack. It is the first method known to generate adversarial image samples that are extremely close to their innocuous counterparts.\\

This attack tries to minimize the distance between the original sample and a modified, misclassified sample, until it is close enough to be indistinguishable from the original. It is a way for an attacker to optimize, for different values of a perturbation norm constraint, a loss-maximization problem until it finds an optimally undistinguishable sample.\\

\subsubsection{FGSM}The Fast Gradient Sign Method or \emph{FGSM} tries to modify a sample by adding or substracting a small value epsilon for each dimension of the input. More formally, the attacker generates a perturbation vector $\lambda$ such that :

$$ \lambda = \epsilon \ \text{sgn}(\nabla_{x}\mathbf{L}(f(x),y)) $$
 
In untargeted mode, the attacker adds a small noise over the input, adding $\pm\epsilon$ to every dimension of the input in order to increase the error, choosing the sign accordingly.\\

In targeted mode, the attacker generates the highest output for the desired class, and thus the lowest error relatively to the adversary's target. The vector thus becomes :\\

$$ \lambda = -\epsilon \ \text{sgn} (\nabla_{x}\mathbf{L}(f(x),y_{adv})) $$

Where $\mathbf{L}$ is the network's loss function.

In both cases, a clipping is made to make sure the values stay in their original range. The sum, along hundreds of dimensions, of small input/output error gradients can result in a large perturbation. Thus, an attacker can easily \emph{push} the sample through a decision boundary and trick the system into a misclassification.\\

This attack is easy to compute and, for some unprotected classifiers, creates large errors. Computing the gradient values requires a White-Box access to the network.
\paragraph{The FGSM on the field} As the FGSM is a \emph{White-Box} attack, the attacker needs access to privileged information. Using the information, the attacker takes a picture of \emph{any check}. The original sample is likely to be classified as the original check. The attacker then computes, for each color channel of each pixel, whether they add or substract, $\epsilon$ to each value, by looking at the sign of the loss' gradient with respect to every dimension of its input.\\


The $\epsilon$ value is chosen according to a tradeoff: the lower the value, the closer the picture will be to the original one, thus making it harder for a system administrator to see whether there was an attack or not. On the other hand, the larger $\epsilon$ is, the higher the chances are for the attack to work.

\subsubsection{Carlini \& Wagner attack}
Carlini \& Wagner \cite{Carlini2017CWATTACK} have proposed an attack method based on a custom gradient descent with different candidate losses in order to \emph{learn the perturbation} the same way the network learns.

One cannot use the optimization problem in (\ref{opti}) to find adversarial samples directly. Rather, loss functions point toward the main direction of attack while perturbation norm stays a constraint that can be hard (strict constraint) or loose \cite{Carlini2017CWATTACK}. In the latter, going further from the constraint adds to the loss. 

The loss used by the authors in \cite{Carlini2017CWATTACK} is a composite based on the original network's loss. They try different losses, linked to different attack behaviours.\\

These loss functions force the adversary to learn a perturbation that maximizes the confidence of the system on its erroneous decision. When using a loose constraint, they smooth out the clipping process. As the constraints become differentiable, the authors allow gradient-based methods to converge into a local optimum for their problem.

\paragraph{Carlini \& Wagner attacks on the field}
An attacker needs a \textbf{white-box} access to the classifier. They would use a gradient descent method, incorporating one of the losses proposed by the authors. This sample generation algorithm obeys two constraints: the first crushes the perturbation to be zero, but the other pulls it towards a misclassification. Through a weighted sum of the constraints, and given enough computing power, the same kind of optimization that trained the neural network is used to create an adversarial sample.

\subsubsection{DeepFool}
\citet{moosavideepfool_2015} used an entirely different approach to make their attacks : they considered that any classifier has locally-linear boundaries, allowing the local use of hyperplanes. The decision hyperplane is such that if one moves parallel to it, they will never cross the boundary and thus will never find an adversarial example. By taking the orthgonal direction, one can make sure that they have the \emph{fastest} way to burst out of the decision boundary. In order to compensate for the linearity hypothesis, the author use an iterative method to keep these approximations within a close radius.

\paragraph{DeepFool on the field}
A \emph{White-Box} attacker tries to output a wide number of photos triggering the \emph{same} classification confidence for \emph{every} different class. This would yield a local topology of the multi-dimensional geometry of the decision boundary. They could thus identify the closest hyperplane-boundary to their current point. This equivalent naive method would lead to a gigantic amount of computation. The authors use an algorithm based on a per-boundary geometry to quickly find the nearest way out of the current class, thus triggering a misclassification.\\

We will give a practical explanation of DeepFool. The attacker can compute at every step the set of linear rules that makes them be in the class they want to escape, and make a small step away from the closest combination of those rules. They do so iteratively, to account for the global non-linearity of DNN.

\subsubsection{Surrogate Black-Box Models}
So far, every attack required a \emph{White-Box} access to the system. But in 2016, \citet{papernot_practical_2016} used the fact that Adversarial Examples can transfer from one Deep Learning Model to another\cite{szegedy_intriguing_2013} to create \emph{Black-Box} attacks that require \emph{no} access to the network's parameters. They train an approximation of the target model in order to create a \emph{White-Box surrogate}. They can then use the transferability property of the adversarial examples: adversarial samples that fool the approximated model have a high probability of fooling the target model. Some White-Box attacks transfer better than others: for example, FGSM attacks transfer well to different classifiers.

\paragraph{Surrogate Black-Box Models on the field} 
If an attacker has no access to the model other than the \emph{check reading device}, they can first gather a small dataset of handwritten digits, and train a model on them. Through jacobian augmentation\cite{papernot_practical_2016}, the attacker will then distort the digits it has in the direction of the estimated boundaries, and re-submit to the oracle for evaluation: through this process, the surrogate neural network will get a low-performance approximation of the model. This approximate model will be vulnerable to some adversarial examples. The transferability property make them likely to also fool the target model. 

\section{Adversarial AI Defenses}
For a systematic review of defenses, we refer the reader to \citep{akhtar_threat_2018} for their work in the subset of attacks on images.
\subsection{Defenses strategies and our focus}

A defender protecting a DNN can use three different strategies in order to increase the model's robustness to adversarial examples.

\paragraph{Detection Strategy} Defenders can use a \textbf{Detection} algorithm, above or within the network, in order to detect the adversial nature of a submitted sample \citep{
lu_safetynet_2017,
xu_feature_2018,
hendrycks_early_2016,
metzen_detecting_2017,
feinman_detecting_2017,
li_adversarial_2016,
grosse_statistical_2017}. If the sample is detected as adversarial, it is rejected and given to a human reviewer to get its actual meaning.

\paragraph{Reforming Strategy} Defenders can use a \textbf{Reforming} system. In this paradigm, the sample is transformed in a way that modifies the sample's numeric values while preserving its semantics. JPEG Compression, Bit-Depth Reduction and Error Diffusion, are valid examples of such transformations. The Reforming system can also be learned through models that re-create their inputs \cite{meng_magnet_2017,Luo2015FoveationbasedMA,Prakash2018DeflectingAA,2017arXiv170502900D,guo_countering_2017}. The defender thus expects that, by adding error or removing information, they can remove some or all the adversarial perturbations. This is equivalent to a form of noise reduction.

\paragraph{Regularization Strategy} The last defense mechanism one can use is based on the \textbf{Regularization} strategy. In this paradigm, the behaviour of the original system is regularized during a training in order to modify the way it handles samples that are out of the classical distribution. This is equivalent to a form of patching: by changing the program's behaviour, the defender makes an attack ineffective against it.

We present below a parallel with SQL Injections, a well-known problem in information security.\\
\begin{center}

\begin{tabular}{|c|c|}
\hline 
AI Defenses & SQLI Defenses \\ 
\hline 
Detection & Detect special characters \\ 
\hline 
Reforming & Replace special characters \\ 
\hline 
Regularization & Parametric statements \\ 
\hline 
\end{tabular} 
\end{center}

We will focus on the \textbf{Regularization} strategy for the following reasons:

\begin{itemize}
\item A \textbf{Detection} system creates an opportunity for a Denial-of-Service (DoS): the attacker could slightly transform the input of a user through a Man-In-The-Middle attack and would get a consistent amount of rejection, while keeping the feedback sample undistinguishable from the original one.
\item A \textbf{Learnable} \textbf{Reforming} system allows an attacker to exploit another neural network. If succesfully attacked, the reformer would regenerate a sample that is close to the target class chosen by the attacker.
\item A \textbf{Non-Learnable} \textbf{Reforming} system creates a static defender that an adversary can progressively learn to bypass.
\item \textbf{Regularization} strategies have the advantage of presenting a smaller attack surface, as there would be no other algorithm than the one used for inference. They trade this advantage for the requirement of retraining the whole system, opening themselves to poisoning attacks and creating a high upfront cost for the defender.
\end{itemize}

As we consider that DNN need to be trained at least once, the \textbf{Regularization} strategy will be preferred. We highlight however that other strategies are less ressource-intensive as they can often be implemented without retraining the model.

\subsection{Current Regularization Techniques}
\subsubsection*{Adversarial Training}
One of the first ideas introduced in the defender's arsenal, the Adversarial Training\cite{szegedy_intriguing_2013,goodfellow_explaining_2014} is based on generating adversarial samples through white-box attacks in order to add them to the existing training set. By triggering the error, and backpropagating it through the network, the system learns to resist an attack the same way it learns the task at hand.\\

From an information security perspective, this method has the disadvantage of defending the network against \emph{known} attacks only. There is no guarantee that adversarial training protects against attacks other than the attacks the system has been trained on. Moreover, this defense can hinder the network's performance. \citet{madry_towards_2017} address this concern by introducing the idea of an optimal first-order adversary that could subsum every attacker with the same order constraints.\\

With these theoretical guarantees, this approach thus transforms offensive strategies into defensive strategies. This approach searches for optimal attacks to train the network upon. If successfully trained on strong adversarial examples, the system can have security guarantees.\\

\subsubsection*{Gradient Masking/Gradient Shattering}
As all current techniques employ some form of gradient computation, one of the ideas introduced was to reduce the gradient or otherwise hinder the attacker's access to the gradient. \citet{gu_towards_2014} directly penalize the network to forcefully lower its gradients through a regularization, while \citet{papernot_distillation_2015} use distillation. Distillation is a process where a first network is trained on one-hot labels $(\delta_{i=class})_{i\in[1,k]}$ where $k$ is the number of classes. Then, the first network is discarded, and the predictions of the trained model are used as the new labels. This teacher network smoothes-out the labels. A second network is then trained on these smooth labels, thus considerably lowering the gradient values. The authors claim that the reduction factor is above 1000.\\

Black-Box surrogate attacks \citep{papernot_practical_2016} have bypassed this \textit{gradient masking} defense: these attacks do \emph{not} need any gradient. Thus, the only way defensive distillation can impact these gardientless attacks is by regularizing the network's behaviour. Unfortunately, even if a modified and extended defensive distillation has been able to resist some attacks \cite{papernot_extending_2017}, some attacks \citep{Carlini2017CWATTACK} can still generate adversarial samples bypassing this defense.\\

\citet{Nayebi_bio_neural} have used a regularization function rewarding using the saturation values of activations. Network activations that are far from a zero-gradient point are penalized through a regularization. \\

Various similar methods create a non-differentiable variant of their model \emph{at test time} to deny any access to the gradient: \citet{athalye_obfuscated_2018} have systematically reviewed --- and bypassed --- these so-called "gradient shattering" techniques.\\

\citet{ross_improving_2017} add another regularization term to the loss. Rather than only regularizing the parameters' values during the training, they also regularize the value of the input gradient through \emph{double backpropagation}\cite{Drucker_doublebackprop}. This gradient reflects how much each variation of the input can change the class. By diminishing this value, the defender forces any attacker to create larger perturbations. The authors claim that successful adversarial samples against their system have had their meanings modified, and thus cannot be deemed adversarial.
 
\subsection{Other defensive techniques}

\subsubsection{Feature Squeezing}
\citet{xu_feature_2018} have shown that, by reducing the attacker's sample space by using semantic-preserving compression / filtering (eg. going from greyscale to Black and White images), one can detect adversarial samples. We do so using the fact that legitimate inputs often create agreeing predictions when inferred on through different compressions. For adversarial samples, these predictions are different.

\subsubsection{Thermometer Encoding} Discovered by \citet{buckman_thermometer_2018}, this method discretizes the values of the dataset according to arbitrary thresholds. As an example, if we have 4 tresholds at $(80,60,40,20)$ and values between 0 and 100, $42$ will be encoded as $(0,0,1,1)$, and $77$ as $(0,1,1,1)$

\section{Motivations for Spartan Networks}
\subsection{Discretization}

When given labelled data, a classifier is usually given dimensonally-large inputs. Input dimensionalities can vary from several hundred to a milion. However, label space dimensionality ranges from 2 to a hundred. This means that the input space is orders of magnitude larger than the output space.\\

For example, the CAL10K music dataset\cite{tingle2010exploring} has a compressed size of 2.1GB, but the size of the annotations file is 10MB. For the images dataset, the standard MNIST handwritten digits dataset\cite{MNIST} has 10 different possible annotations and thus can be stored on four bits. The digital image needs 6272 bits to be stored. \\


DNN can succesfully create a reliable approximator that can classify correctly and reliably given an appropriate amount of labelled data in standard settings. Adversarial examples show that there are pockets of inputs present in the high-dimensional space that are missclassified. Adversarial training provides a way to change the function approximated by neural networks, but it cannot guarantee that it reliably "patched" the system. Even if augmented, the number of training datapoints is indeed negligible in comparison to the number of possible inputs for a given class. High-dimensionality creates a disadvantage for the defender. \\

Given an adequate $\epsilon$ (\ref{opti}) the perturbation is not perceptible to a human observer, but successfully tricks the network into giving a wrong answer. This means that:
\begin{itemize}
\item the input space of encoded samples has a local density greater than human distinguishability power.
\item the network behaviour is extremely sensitive to small changes.
\end{itemize}

In neural layers using a ReLU or ReLU-equivalent activation function, stimulus perception is almost always near linear or near multilinear. Thus, any small variation of the stimulus can be propagated through the neural network using the fact that all operations are differentiable. The ReLU is made to be equal to the identity function when recieving a positive input, meaning that the ReLU is a linear function when activated.\\

As an example, human observers cannot distinguish colors between $\{61, 175, 250\}_{\text{RGB}}$ and $\{61, 170, 250\}_{\text{RGB}}$, but can distinguish cyan from blue. The DNN however can distinguish the former extremely well.\\

The attacker, by modifying the value of parts of the input by a small amount, can also make the value of activation function vary by a proportional amount. This perturbation will be propagated to all affine combinations including the perturbed inputs. As the propagation goes forward, the perturbation, if cleverly crafted by the attacker, spreads and its amplitude grows within the network, using the weights of the network to trick and force the network to output another value. This exploitation can be done for any network having a succession of locally-linear activation functions, through the Taylor expansion of any differentiable function.

Various authors have stated \cite{abbasi_robustness_2017,buckman_thermometer_2018} that excessive linearity seems to be a point of vulnerability of neural networks.

From \citet{xu_feature_2018}, we learned that squeezing features could sometimes prevent adversarial perturbations from being effective. \citet{buckman_thermometer_2018} extend this idea with the Thermometer Encoding. We extend on the Thermometer Encoding strategy by learning the tresholds that are useful to correctly do the task while using the minimal number of information.\\

Hence our hypotheses are as follows:
\begin{itemize}
\item Current neural networks exhibit locally-linear properties that can be used to slide the sample. classification towards another class.
\item The dimensionality is one of the factors allowing attackers to find a good attack vector.
\end{itemize}

We summarize all the problems discussed above in Fig.~\ref{fig:DNNattack}. A Deep Feedforward Neural Network has a very large input space for a small output space and excessive linearity allows the attacker to explore the space and find an adversarial examples.



\begin{center}
\begin{figure}
\tikzset{%
  every neuron/.style={
    circle,
    draw,
    minimum size=1cm
  },
  neuron missing/.style={
    draw=none, 
    scale=4,
    text height=0.333cm,
    execute at begin node=\color{black}$\vdots$
  },
}

\begin{tikzpicture}[x=1.5cm, y=1.5cm, >=stealth]

\foreach \m/\l [count=\y] in {1,2,3,4,5,6}
  \node [every neuron/.try, neuron \m/.try] (input-\m) at (\y,2) {$I_\m$};

\foreach \m [count=\y] in {1,2,3,4,5,6}
  \node [every neuron/.try, neuron \m/.try ] (hidden-\m) at (\y,4) {ReLU};
  
\foreach \m [count=\y] in {1,2,3,4,5,6}
  \node [every neuron/.try, neuron \m/.try ] (hidden2-\m) at (\y,6) {ReLU};

\foreach \m [count=\y] in {3,4}
  \node [every neuron/.try, neuron \m/.try ] (output-\m) at (\m,8) {SMax};



\foreach \l [count=\i] in {3,4}
  \draw [->] (output-\l) -- ++(0,1);
\foreach \l [count=\i] in {3,4}
  \node [above right] at (output-\l.north){$O_\i$};

\foreach \i in {1,2,3,4,5,6}
  \foreach \j in {1,2,3,4,6}
    \draw [->] (input-\i) -- (hidden-\j);
    
  \foreach \i in {1,2,3,4,5}
    \draw [->] (input-\i) -- (hidden-5);
    
\draw [->, line width=1mm, color=red] (input-6) -- (hidden-5);

\foreach \i in {1,...,6}
  \foreach \j in {1,...,6}
    \draw [->] (hidden-\i) -- (hidden2-\j);
    
\draw [->, line width=1mm, color=red] (hidden-5) -- (hidden2-2);

\foreach \i in {1,...,6}
  \foreach \j in {3,4}
    \draw [->] (hidden2-\i) -- (output-\j);
    
\draw [->, line width=1mm, color=blue] (hidden2-2) -- (output-3);
    
\draw [->, line width=1mm, color=red] (hidden2-2) -- (output-4);


\end{tikzpicture}

\caption{A Deep Feedforward Neural Network. If the ReLU or any locally-linear activation function is used, an attacker can find, out of all the possible weights, the highest weights bending prediction towards misclassification. Red means an increase, blue a decrease. By increasing the value of a part of the input, the attacker can linearly manipulate part of the calculation and decrease the current probability of the current class and increase the probabilty of another}
\label{fig:DNNattack}
\end{figure}
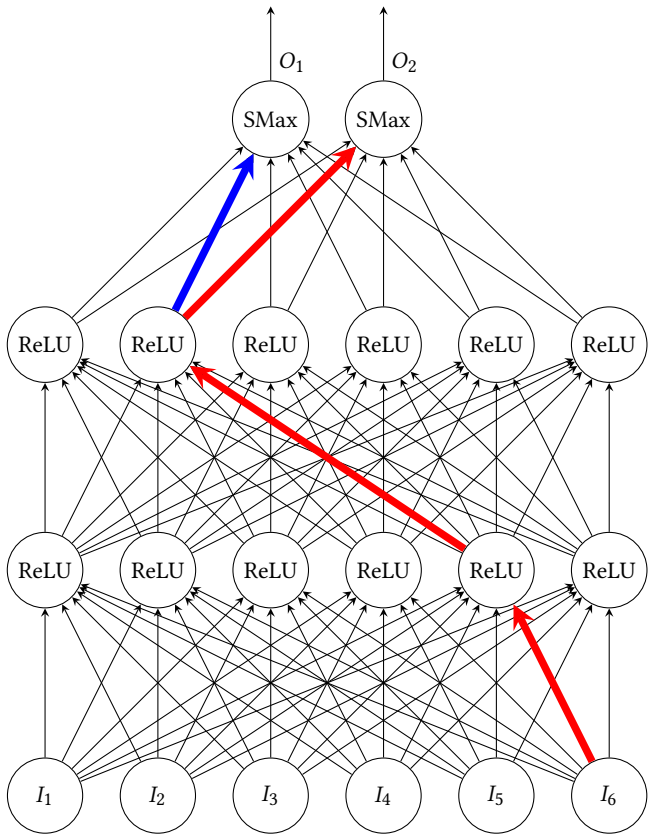

\end{center}


\begin{center}

\begin{figure}

\includegraphics[width=0.2\textwidth]{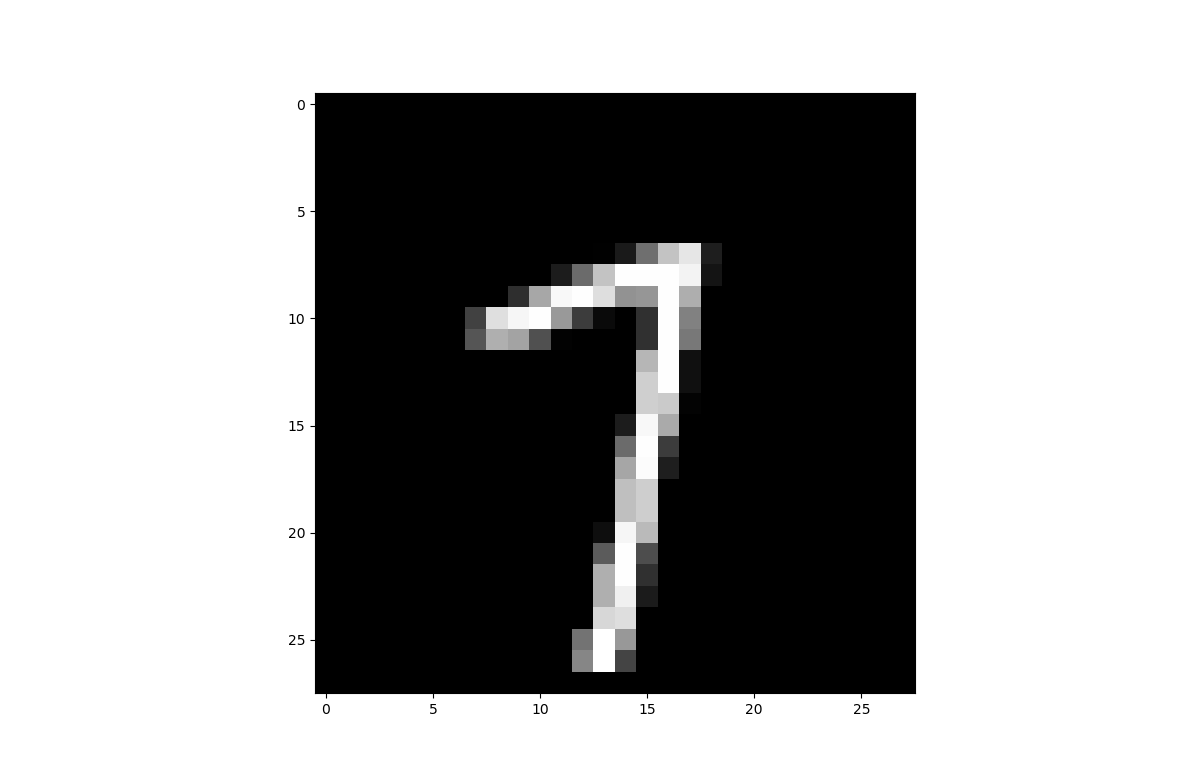}
\includegraphics[width=0.2\textwidth]{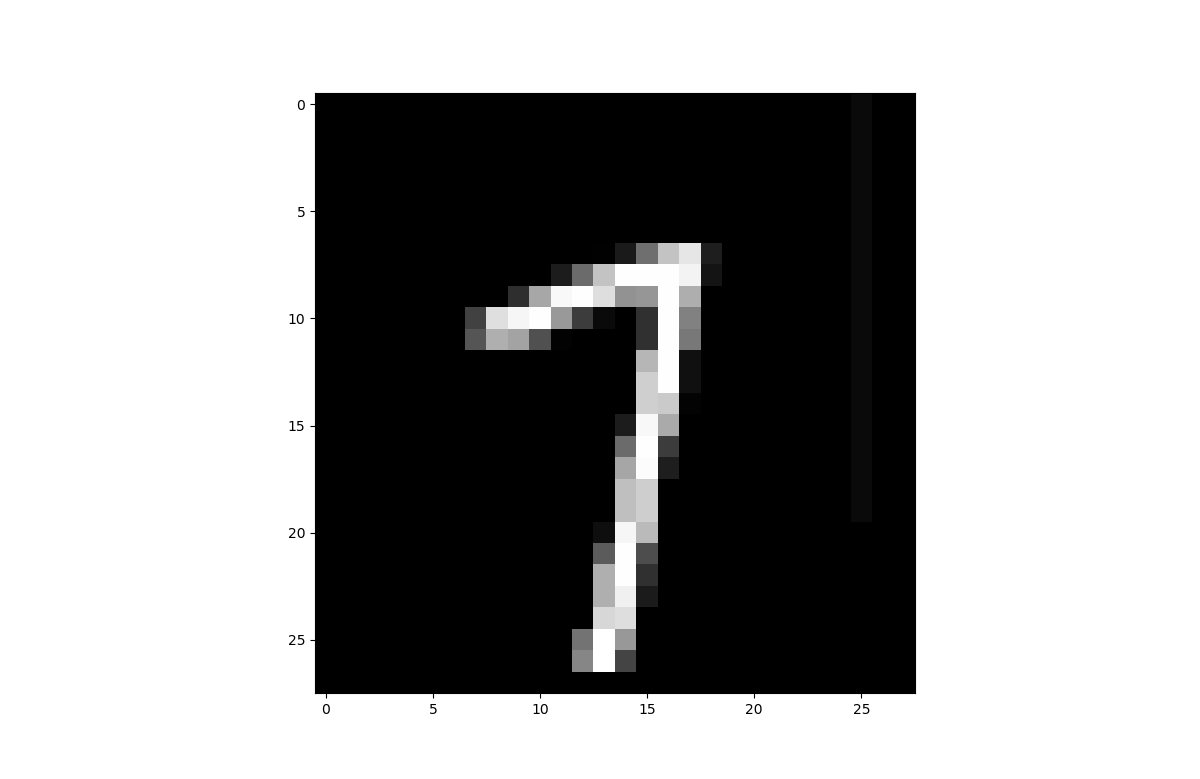}
\includegraphics[width=0.2\textwidth]{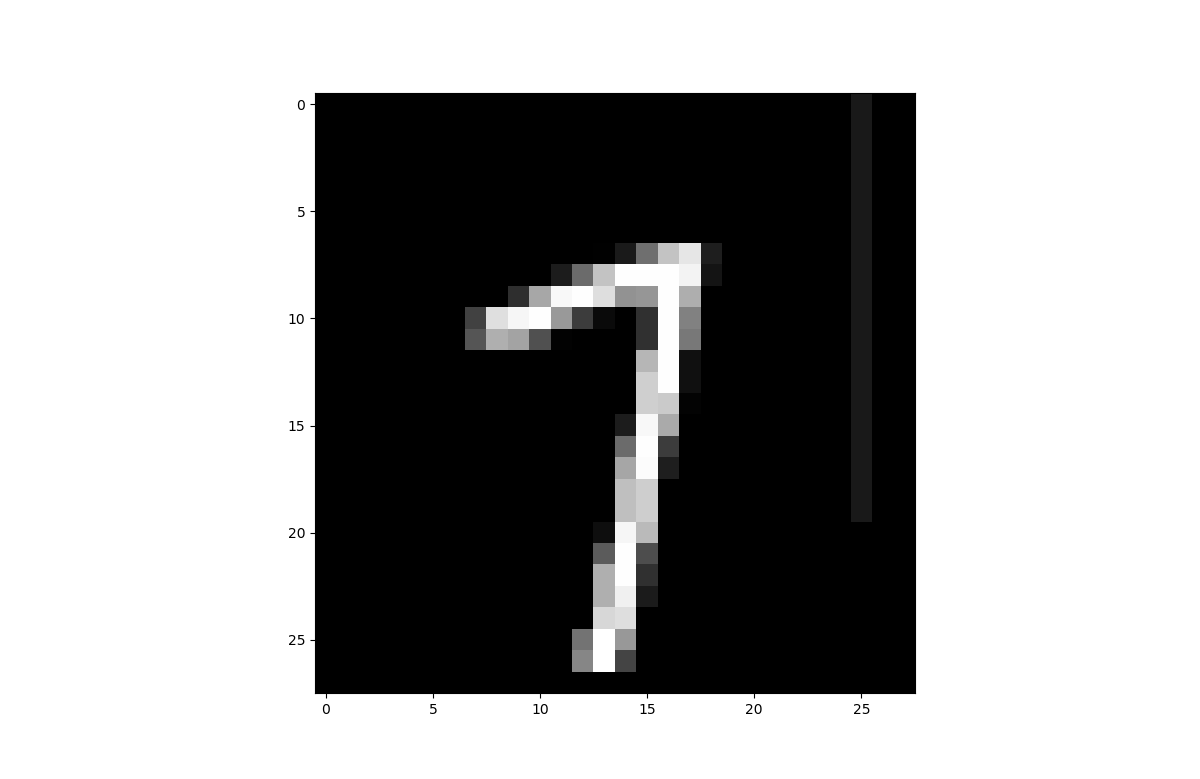}
\includegraphics[width=0.2\textwidth]{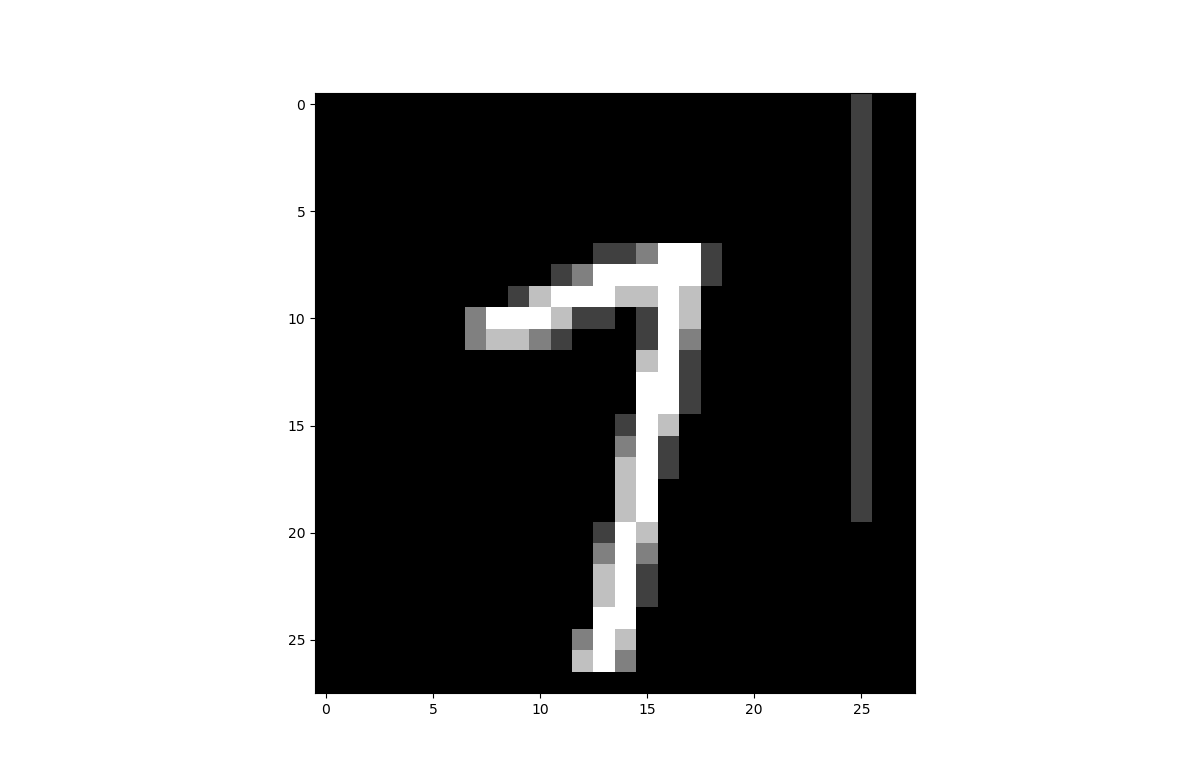}
\caption{In reading order: a non-perturbed image, an image with a little, 5\% white, vertical bar next to the number, the 10\% version, and a thresholded version, with 3 equidistant thresholds, effectively putting the perturbation at 25\%. The perturbation becomes much easier to see, while the image keeps its meaning}
\label{fig:perturb}
\end{figure}
\end{center}

The human response to color seems to be non-linear, as seen in Fig~\ref{fig:perturb}. The little perturbation is not seen, and double this perturbation is. The tresholding does not destroy any semantics, as one can recognize the digit.\\

When stimuli values are discretized, small changes to any stimulus must go past the next threshold to create any perturbation. The attacker will thus have to push the values further than it normally would in order to put a valid perturbation, that would be visible to an observer. Two samples on the same threshold are indeed considered as the same to the system. The work of \citet{buckman_thermometer_2018} is using this idea.\\

The non-linearity of discretization as well as its saturation properties could restrain an attacker from propagating its perturbations throughout any DNN.\\

Previous attemps failed at training with staircase-like functions, that are inherently non-differentiable, as the 'derivative' of these functions is a sequence of impulses.


\subsection{Binary Encoding}
As seen previously in the works of \citet{buckman_thermometer_2018}, one can use a vector-of-bits encoding to make a network more robust against adversarial perturbations. Combined with the work of \citet{Nayebi_bio_neural}, we hypothesize that robust learning in DNN can be achieved by:
\begin{itemize}
\item making their behaviour as non-linear and saturated as possible;
\item reducing the attack space for the attacker by squeezing it;
\item creating a binary array instead of float activation values.
\end{itemize}
We aim at learning an optimal way to achieve this through backpropagation, by creating new layers of neurons. These neurons will learn this behaviour by themselves, however we make sure that the following requirements are met:
\begin{itemize}
\item The performance on the test dataset must stay relatively close to an unprotected counterpart.
\item We must find a way to learn a highly non-differentiable filtering that can interface with backpropagation.
\end{itemize}

\begin{figure}

\tikzset{%
  every neuron/.style={
    circle,
    draw,
    minimum size=1cm
  },
  neuron missing/.style={
    draw=none, 
    scale=4,
    text height=0.333cm,
    execute at begin node=\color{black}$\vdots$
  },
}

\begin{tikzpicture}[x=1.5cm, y=1.5cm, >=stealth]

\foreach \m/\l [count=\y] in {1,2,3,4,5,6}
  \node [every neuron/.try, neuron \m/.try] (input-\m) at (\y,2) {$I_\m$};

\foreach \m [count=\y] in {1,2,3,4,5,6}
  \node [every neuron/.try, neuron \m/.try ] (hidden-\m) at (\y,4) {DU};
  
\foreach \m [count=\y] in {1,2,3,4,5,6}
  \node [every neuron/.try, neuron \m/.try ] (hidden2-\m) at (\y,6) {ReLU};

\foreach \m [count=\y] in {3,4}
  \node [every neuron/.try, neuron \m/.try ] (output-\m) at (\m,8) {SMax};



\foreach \l [count=\i] in {3,4}
  \draw [->] (output-\l) -- ++(0,1);
\foreach \l [count=\i] in {3,4}
  \node [above right] at (output-\l.north){$O_\i$};

\foreach \i in {1,2,3,4,5,6}
  \foreach \j in {1,2,3,4,6}
    \draw [->] (input-\i) -- (hidden-\j);
    
  \foreach \i in {1,2,3,4,5}
    \draw [->] (input-\i) -- (hidden-5);
    
\draw [->, line width=1mm, color=red] (input-6) -- (hidden-5);

\foreach \i in {1,...,6}
  \foreach \j in {1,...,6}
    \draw [->] (hidden-\i) -- (hidden2-\j);
    
\draw [->, line width=0.7mm, dashed, color=red!55!gray] (hidden-5) -- (hidden2-2);

\foreach \i in {1,...,6}
  \foreach \j in {3,4}
    \draw [->] (hidden2-\i) -- (output-\j);

\draw [->, line width=0.8mm, color=blue!27!gray, dashed] (hidden2-2) -- (output-3);
\draw [->, line width=0.8mm, color=red!27!gray, dashed] (hidden2-2) -- (output-4);


\end{tikzpicture}
\caption{By breaking the linearity and outputting only discrete values, we make sure that an attacker cannot use gradient descent to find an adversarial example. DU stands for Discrete Unit.}
\label{fig:DNNheaviside}
\end{figure}
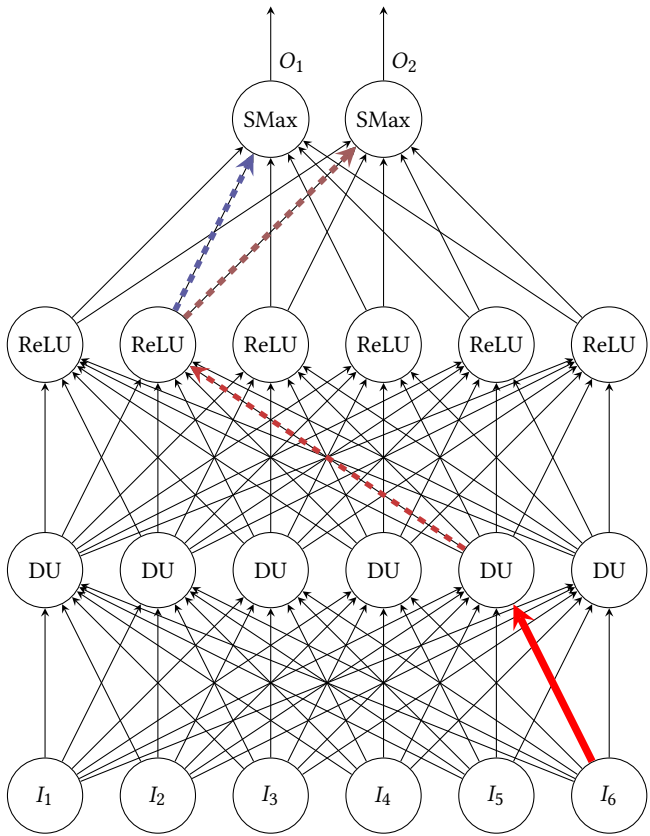

We thus create a processing layer, that we called filtering layer, whose behaviour is made to be extremely non-linear. This layer has a discrete, low-cardinality value range. We can train it through backpropagation. We modify the DNN of figure~\ref{fig:DNNattack} into the DNN of figure~\ref{fig:DNNheaviside} with these ideas. Note that \citet{courbariaux_binarized_2016} have already proposed restricted-cardinality activation functions.\\

The processing would use the Heaviside step function, where the activation function's definition range is orders of magnitude larger than the output range. In this case, 9 orders of magnitude for the Heaviside, instead of less than 1 for the ReLU. As we get less information-bits for each neuron, we destroy more information. This process will be called \emph{data-starving}.\\

To better understand the interest of data-starving, we emphasize that most activation functions are encoded on from 16 to 64 bits, most commonly 32. 32-bit-float activation functions like the 32-ReLU have $2^{31}$ possible values, as only positive values are different from one another. While these activation functions show a high number of possible states, training set cardinalities are orders of magitude lower than the possible number of states. Thus, unexplored areas in those activation functions can be exploited by an adversary to find adversarial examples. The defender can thus try to reduce the amount of states in each layer by various methods. These methods include using lower weights, or adding a saturation value. This idea was also exposed in the work of \citet{xu_feature_2018}.\\

We decide to reduce the attacker's space as much as possible, by taking an activation function that outputs only \emph{two} possible values. We will thus use the Heaviside step function as an activation function in the filtering Layer.

\subsection{Spartan Training}

During the training, the filtering layers data-starve the network because there are few possible output values out of them. Moreover, these layers are regularized by the amount of signal they let through. To do this, we use a $\mathcal{L}_1$ loss on the activations of the network. Thus, this added loss is proportional to the amount of signal that the system let through. In addition to this loss, the network uses a training loss based on the error, which is standard for training DNN. The training loss plays directly against the loss of the filtering layer.\\ 

To lower the \textbf{training loss}, the system needs to get information that increases the \textbf{filtering loss}. The only way to reduce the \textbf{training loss} is by destroying information that is likely to increase the \textbf{filtering loss}. We thus have two competitors within the network fighting to reduce their losses.\\

The self-adversarial behaviour of the network shall allow the system to harden itself during this seemingly \emph{Spartan Training}\footnote{The ancient Spartan Training was known to be extremely harsh.}, by allowing it to reduce the value ranges within the network activations. The attacker will either have to generate a high-amplitude adversarial noise, or stay in the regulated input space, more restricted, requiring more computing power, and thus yielding a higher cost for the attacker.\\

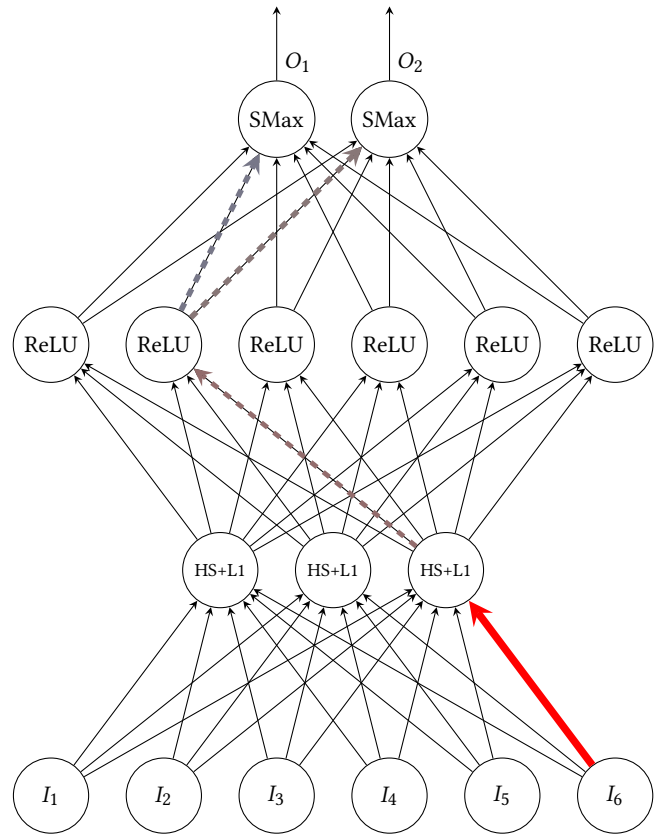
\begin{figure}

\tikzset{%
  every neuron/.style={
    circle,
    draw,
    minimum size=1cm
  },
  neuron missing/.style={
    draw=none, 
    scale=4,
    text height=0.333cm,
    execute at begin node=\color{black}$\vdots$
  },
}

\begin{tikzpicture}[x=1.5cm, y=1.5cm, >=stealth]

\foreach \m/\l [count=\y] in {1,2,3,4,5,6}
  \node [every neuron/.try, neuron \m/.try] (input-\m) at (\y,2) {$I_\m$};

\foreach \m [count=\y] in {3,4,5}
  \node [every neuron/.try, neuron \m/.try ] (hidden-\m) at (\m-0.5,4) {\footnotesize HS+L1};
  
\foreach \m [count=\y] in {1,2,3,4,5,6}
  \node [every neuron/.try, neuron \m/.try ] (hidden2-\m) at (\y,6) {ReLU};

\foreach \m [count=\y] in {3,4}
  \node [every neuron/.try, neuron \m/.try ] (output-\m) at (\m,8) {SMax};



\foreach \l [count=\i] in {3,4}
  \draw [->] (output-\l) -- ++(0,1);
\foreach \l [count=\i] in {3,4}
  \node [above right] at (output-\l.north){$O_\i$};

\foreach \i in {1,...,6}
  \foreach \j in {3,4}
    \draw [->] (input-\i) -- (hidden-\j);
    
  \foreach \i in {1,2,3,4,5}
    \draw [->] (input-\i) -- (hidden-5);
    
\draw [->, line width=1mm, color=red] (input-6) -- (hidden-5);

\foreach \i in {3,4,5}
  \foreach \j in {1,...,6}
    \draw [->] (hidden-\i) -- (hidden2-\j);
    
\draw [->, line width=0.65mm, dashed, color=red!15!gray] (hidden-5) -- (hidden2-2);

\foreach \i in {1,...,6}
  \foreach \j in {3,4}
    \draw [->] (hidden2-\i) -- (output-\j);

\draw [->, line width=0.75mm, color=blue!7!gray, dashed] (hidden2-2) -- (output-3);
\draw [->, line width=0.75mm, color=red!7!gray, dashed] (hidden2-2) -- (output-4);


\end{tikzpicture}

\caption{The Spartan version of the DNN: the first layer is severely squeezing the dimensionality of the input, and outputs binary values only. Furthermore, the activation of each neuron is added to the global loss of the neural network (L1), forcing this layer to output as few signal as possible.}
\label{fig:DNNspartan}
\end{figure}

We define the network composed of a data-starving layer connected to a neural network and trained with a composite loss a \emph{Spartan Network}. To understand the Spartan Training and the filtering layer effect, we transform the Discrete-DNN of Fig.~\ref{fig:DNNheaviside} into the Spartan DNN of figure~\ref{fig:DNNspartan}.

%
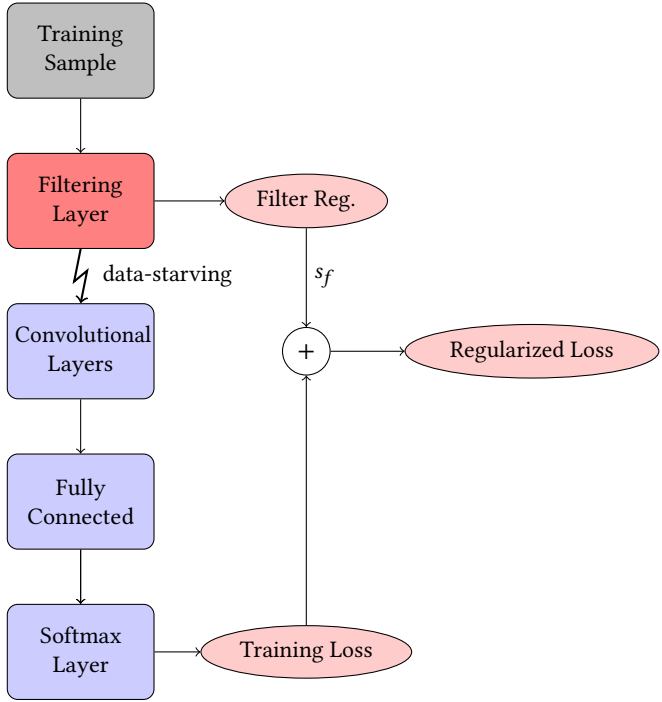
\begin{figure}[h]
\tikzstyle{decision} = [diamond, draw, fill=blue!20, 
    text width=4.5em, text badly centered, node distance=3cm, inner sep=0pt]
\tikzstyle{block} = [rectangle, draw, fill=blue!20, 
    text width=5.5em, text centered, rounded corners, minimum height=4em]
\tikzstyle{line} = [draw, -latex']
\tikzstyle{cloud} = [draw, ellipse,fill=red!20, node distance=3cm,
    minimum height=2em]
\tikzstyle{circlep} = [draw, circle, text centered, node distance=3cm,
    minimum height=2em]

\pgfdeclaredecoration{lightning bolt}{draw}{
\state{draw}[width=\pgfdecoratedpathlength]{
  \pgfpathmoveto{\pgfpointorigin}
  \pgfpathlineto{\pgfpoint{\pgfdecoratedpathlength*0.6}%
    {-\pgfdecoratedpathlength*.12}}%
  \pgfpathlineto{\pgfpoint{\pgfdecoratedpathlength*0.4}{\pgfdecoratedpathlength*.12}}%
  \pgfpathlineto{\pgfpoint{\pgfdecoratedpathlength}{0pt}}%
}%
}

\centering
\begin{tikzpicture}[node distance = 2cm, auto]
    \node [block, fill=gray!50] (ts) {Training Sample};
    \node [block, below of=ts, fill=red!50] (fl) {Filtering Layer};
%
    \node [cloud, right of=fl] (lfl) {Filter Reg.};
    \node [block, below of=fl] (nn) {Convolutional Layers};
    \node [circlep, right of=nn ] (sm) {\Large $+$};
    \node [cloud, right of=sm] (tl) {Regularized Loss};
    \node [block, below of=nn] (fc) {Fully Connected};
    \node [block, below of=fc] (ol) {Softmax Layer};
    \node [cloud, right of=ol] (lil) {Training Loss};
    
    \draw [->] (ts) -- (fl);
    \draw [->] (fl) -- (lfl);
    \draw [->,decorate,decoration={lightning bolt}, thick] (fl) -- node [right=5pt] {data-starving} (nn);
    \draw [->] (nn) -- (fc);
    \draw [->] (fc) -- (ol);
    \draw [->] (ol) -- (lil);
    \draw [->] (fc) -- (ol);
    \draw [->] (lfl) -- node[right]{$s_{f}$} (sm);
    \draw [->] (lil) -- (sm);
    \draw [->] (sm) -- (tl);
    
\end{tikzpicture}
  \caption{Example structure of a Convolutional Spartan Network for images, $s_{f}$ is the \emph{scaling factor}, the weight of the filtering regularization relatively to the training loss}
\label{fig:SpartanStruct}
\end{figure}
\section{Spartan Network structures}
In this section, we create three filtering layers and explore the three different Spartan Networks using them.

\subsection{Composite activation function}
As stated before, we use the Heaviside step function on the forward propagation. This activation function's gradient is zero where it is defined. The filtering layer thus uses a synthetic gradient in order to use backpropagation through this zero gradient.

The Heaviside step function's definition is:
\begin{equation}\begin{aligned} \label{eqHS}
&H(x) : \mathbb{R} \rightarrow \{0,1\} \\
&H(x) =\left\{\begin{aligned}
&1 \ \text{if}\ x\geq0 \\ 
&0 \ \text{else} 
\end{aligned}
\right.
\end{aligned}
\end{equation}

On the backwards propagation, the function could be, for example, replaced with the arctangent activation function, with the following properties:

\begin{equation} \label{eqAT}
\text{arctan'}(x) = \frac{1}{1+x^{2}}
\end{equation}

Mixing definitions \ref{eqHS} and \ref{eqAT}, we get the HSAT activation function:
\begin{equation} \label{eqHSAT}
\begin{aligned}
&\text{HSAT}(x)= H(x), \\
&\text{HSAT'}(x) \longleftarrow \text{arctan'}(x) = \frac{1}{1+x^{2}}
\end{aligned}
\end{equation}

We use a mnemonic of a non-differentiable (or trivially differentiable) function concatenated with a two-letter mnemonic of the differentiable function whose derivative is used for gradient computing.\\

HSAT will then be the Heaviside-Arctangent activation function, HSID the Heaviside-Identity, Cos-HSAT for a Cosine-Heaviside-Identity function. To simplify, we consider these activation functions to be one activation function, that we will call \emph{composite activation function}.

We have investigated four such functions. In the following, $H$ is the Heaviside step function, and $\Phi$ is the normal distribution used as a function.\\
\begin{center}

\begin{tabular}{|c|c|c|}
\hline 
Forward & Backward & Mnemonic \\ 
\hline\hline
$H$ & $\text{id}: f(x)=x$ & HSID \\ 
\hline 
$H$ & $\arctan$ & HSAT \\ 
\hline 
$H \circ \cos$ & $\arctan\circ\cos$ &Cos-HSAT \\ 
\hline 
$H \circ \cos $& $\Phi\circ\cos$ & Cos-HSND\\ 
\hline 
\end{tabular} 
\end{center}

Feedforward Neural Networks are trained through backpropagation, and, while other weight updates exist, this method has demonstrated its power over the past years. If we were to use backpropagation \emph{as-is} on this filtering layer however, we would be unable to update its weights or biases, as the derivative of the Heaviside step function is zero on all values where the function is differentiable. This is due to the fact that the Heaviside step function is the integral of the Dirac impulse function $\delta$. Thus, no gradient can be used to update the biases without the synthetic gradient. Hence the decision to arbitrarily chose to replace the Heaviside step function's derivative by another function's derivative on the backwards pass.\\

We will test some replacement derivative function candidates stated above in our Experiments section.\\


With the idea of a Forward-Backwards composite activation function, we decouple the forward propagation from the backwards propagation dynamics on this layer, opening the path to use more complex update functions.\\

Note that this separation idea has also been explored in a gradient approximation attack context in the work of \citet{athalye_obfuscated_2018}, under the name \emph{Backward Pass Differentiable Approximation} or BPDA. We differ from this work as the selection of the Backward pass is \emph{arbitrary}, and has no need to be a close approximation of the function. We nonetheless keep the derivative sign to mirror the general behaviour of the original function.


\subsection{Candidate Filtering layers}
Aiming to learn different thermometer encodings \cite{buckman_thermometer_2018} through backpropagation, the filtering layers focus on destroying irrelevant information. We propose and implement candidate layers exhibiting this property.

We will take the number of dimensions of the encoding as an hyperparameter $\beta$. $\beta=4$, for example, will mean that there are four thresholds, and thus five possible values for the encoding.
\subsubsection{Convolution-Filtering}
The simplest way filter information with convolutions is to use the Heaviside step activation function into standard convolutional layers with a $1\times1$ kernel.\\

For every filter, each channel of an input image will be multiplied by a learned parameter and a bias will be added, before going through the Heaviside activation function

This non-differentiable activation function can allow the network to output thermometer encodings, as, with $b$ the bias of the unit and $w$ a weight:

\begin{equation} \label{eqHSactivate}
\text{HS}(wx+b) = 1 \Leftrightarrow wx+b>0 \Leftrightarrow x>-\frac{b}{w} 
\end{equation}

By having various values for $-\frac{b}{w}$ we can create a thermometer encoding created by an activation function and learned through backpropagation, as was our goal. 

\subsubsection{Offset-Filtering}

To prove that the robustness of the model is not convolution-dependent, we can also implement the filtering using a locally connected layer. One can vary the amount of neurons in the layer. The filtering layer is as large as the input. The neurons in the layer have only one connection, and their weights are constrained to be one. Only their biases are learned during training. We effectively create a composition between a binary filter and an image mask learned by the system.

\subsection{Candidate composite activation functions}

\subsubsection{HSID}
The Heaviside-Identity activation function definition is as follows:

\begin{equation}
\begin{aligned}
\text{HSAT}(x)&= H(x), \\
\text{HSAT'}(x) &\longleftarrow \text{id}_{\mathbb{R}}'(x) \\
\text{HSAT'}(x) &\longleftarrow 1 \\
\end{aligned}
\label{eq:HSID}
\end{equation}

This composite activation function is interesting because the backward pass is not an approximation of the foward pass.

\subsubsection{HSAT}
The Heaviside-Arctangent activation function definition is as follows:

\begin{equation}
\begin{aligned}
\text{HSAT}(x)&= H(x), \\
\text{HSAT'}(x) &\longleftarrow \text{arctan'}(x) \\
\text{HSAT'}(x) &\longleftarrow \frac{1}{1+x^{2}}
\end{aligned}
\end{equation}

\subsubsection{HSAT $\circ$ Cosine}
We compose the HSAT function with a cosine in order to get a learnable but square activation function. The Heaviside-Arctangent$\circ$Cosine activation function definition is as follows:

\begin{equation}
\begin{aligned}
\text{HSAT}(\cos(x))&= H(\cos(x)), \\
\text{HSAT'}(cos(x)) &\longleftarrow \text{arctan'}(\cos(x)) \\
\text{HSAT'}(\cos(x)) &\longleftarrow -\frac{\sin(x)}{1+\cos(x)^{2}}
\end{aligned}
\end{equation}

\subsubsection{HSND $\circ$ Cosine}
We modify the previous activation function using the normal distribution as a function for the gradient part.

\begin{equation}
\begin{aligned}
\text{HSND}(\cos(x))&= H(\cos(x)), \\
\text{HSND'}(cos(x)) &\longleftarrow \Phi'(\cos(x)) \\
\text{HSND'}(\cos(x)) &\longleftarrow -x\sin(x)\Phi(x)
\end{aligned}
\end{equation}
We used $\mu=0, \sigma=1$ as the base values for the normal distribution.

While the normal distribution requires more computation, it is mathematically more relevant, as the dirac $\delta$ impulse is the limit of the normal distribution as $\sigma\Rightarrow 0$. We will test different values for $\sigma$ to see whether a synthetic gradient that fits the foward activation function's gradient more closely --- that is, using a $\sigma$ closer to 0 --- yields better results.

\begin{figure}
\begin{tikzpicture}[xscale=0.65,yscale=2,domain=-5:5,samples=400]
    \draw[->] (-5.3,0) -- (5.3,0) node[below] {$x$};
    \draw[->] (0,-0.3) -- (0,1.3) node[left] {$\text{HSAT}(x)$};
   \draw [blue, ultra thick] (-5,0) -- (0,0) -- (0,1) -- (5,1);
    
    \draw[red] plot (\x,{1/(\x*\x+1)});

\end{tikzpicture}

\begin{tikzpicture}[xscale=0.5,yscale=2,domain=-7:7,samples=400]
    \draw[->] (-7.3,0) -- (7.3,0) node[below] {$x$};
    \draw[->] (0,-0.3) -- (0,1.3) node[left] {$\text{HSAT}(cos(x))$};
    \draw[blue] plot (\x,{max(ceil(cos(deg(\x))),0)});
    \draw[red] plot (\x,{-1*sin(deg(\x))/(1+cos(deg(\x))^2)});

\end{tikzpicture}

\begin{tikzpicture}[xscale=0.5,yscale=2,domain=-7:7,samples=400]
    \draw[->] (-7.3,0) -- (7.3,0) node[below] {$x$};
    \draw[->] (0,-0.3) -- (0,1.3) node[left] {$\text{HSND}(cos(x))$};
    \draw[blue] plot (\x,{max(ceil(cos(deg(\x))),0)});
    \draw[red] plot (\x,{-1*sin(deg(\x))*0.39894228*exp(-1*(cos(deg(\x))*cos(deg(\x)))/2)});
    \draw[green, dashed] plot (\x,{-1*sin(deg(\x))/(1+cos(deg(\x))^2)});
    
\end{tikzpicture}

 \caption{Three candidate composite activation functions: the Heaviside-Arctangent ($y=\text{HSAT}(x)$), the Cosine-Heaviside-Arctangent function $y=\text{HSAT}(cos(x))$, and the Cosine-Heaviside-NormalDistribution function $y=\text{HSND}(cos(x))$. The blue plot is the forward activation function, the red plot the backwards derivative. The green plot is the comparison between HSID (\textit{green}) and HSND (\textit{red}) derivatives }
    	\label{fig:PlotsOfActFunctions}
\end{figure}
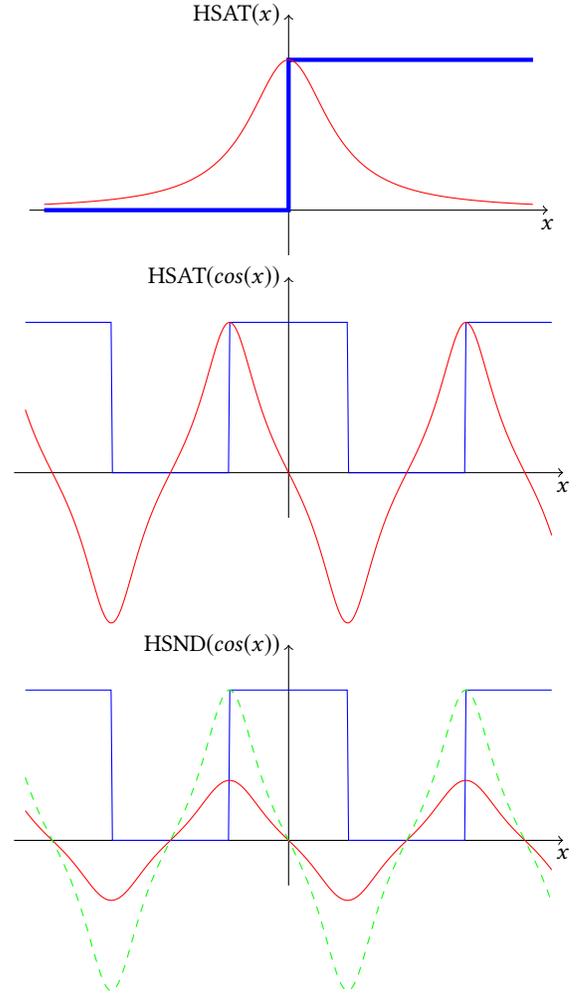

We can see on Fig.~\ref{fig:PlotsOfActFunctions} that the derivative part of the composite activation functions are "smoothed out" variants of their spiky, non-differentiable counterparts.

\subsection{Loss Regularizations of the candidates}
The following section shows the regularizations of the various candidate layers we experimented on.

As seen in Fig.~\ref{fig:SpartanStruct}, Spartan Networks have a scaled loss Regularization that rewards the filtering layer when it data-starves the network if $s_{f}>0$.
\subsubsection{Convolution-Filtering}
A simple $\mathcal{L}_{1}$ activation regularizer is proposed. As the number of activation is reduced, the number of bright, activated pixels diminishes, data-starving the network.

\subsubsection{Offset-Filtering}
The filtering layer adds a term to the loss function to penalize the network if it gives away too much information. We need a loss function that attains its maximum at a half of the function value distrubtion and is 0 at the edges. The entropy function is the best candidate, and we thus define the regularization function as:\\

\begin{equation} \label{eq2}
\begin{split}
L((B_{i})_{i \in \mathbb{N}})&=\sum_{i=1}^{N} B_{i}log(B_{i})\\
B_{i}&=\frac{b_{i}-x^{min}_{i}}{x^{max}_{i}-x^{min}_{i}}
\end{split}
\end{equation}

Where $N$ is the input size, $(x^{max}_{i},x^{min}_{i})$ the maximum and minimum value of the input for the $i^{th}$ input.

$B_{i}$ is the rescaled bias, ranging from 0 to 1. This rescaled bias puts an \emph{a priori} distribution over the dataset.

Note that this particular $B_{i}$ holds when we hypothesize that there is a uniform distribution on the values. In an opposite case, one can create a cut-off based on a cumulative distribution that can be used to rescale the biases. A $B_{i}$ close to $\frac{1}{2}$ will signal that the probability of drawing a sample pixel value above the threshold is the same as below.

\subsection{Data-Starving Behaviours}
\subsubsection{Offset-Filtering}
The filtering layer minimizes its regularization loss if the value of the all rescaled biases of the neurons of the filtering layer are closer to either 0 or 1. Rescaled biases close to 0 or 1 mean neurons change activation close the the minimum or the maximum value. The filtering layer thus forces the network to destroy as much information as it can, due to the behaviour of the entropy function: 
\begin{itemize}
\item When the rescaled bias is at $\frac{1}{2}$, the ranges where $HS(x-B_{i})$ are 0 or 1 are equal and the entropy function is at its maximum: the layer gives more information. We hence penalize the fact that the network thrives on information.
\item When the rescaled bias is close to 0, $HS(x-B_{i})$ is almost always 1: we get no extra information as this feature is bound to be present. Entropy regularization is close to 0
\item When the rescaled bias is close to 1, $HS(x-B_{i})$ is almost always 0. The feature will almost never be present. Entropy regularization is close to 0.
\end{itemize}

\subsubsection{Convolutional-Filtering}
The more the filtering layer activates, the higher the penalty is because of the activation regularization we put on this Filtering Layer. This means that the filters are biased towards higher thresholds, as a higher threshold value will decrease the number of inputs dimensions that cross the threshold, and will thus reduce the activation regularization penalty.

We thus encourage the network to report on a feature only if this feature is deemed extremely relevant, or if the value is extreme. We thus restrict the attacker to a visible attack.

\section{Discussion \& Results}
\subsection{Implementation}
We implement our idea in Keras \cite{chollet2015keras}, using parts of the TensorFlow backend \cite{tensorflow2015-whitepaper} for the implementation of the synthetic gradients. The general structure of the Spartan Network we implemented is as follow (first layer first):\\

\begin{itemize}
\item A Filtering Layer, whose parameters will vary but that will stay on the first layer. (Filtering Layer 1)
\item A Convolutional Layer, 32 filters, $3\times 3$ kernel, no stride, ReLU activation function.
\item Another Convolutional Layer, same parameters. ReLU or Composite activation function. (Optional Filtering Layer 2)
\item A $2\times 2$ Pooling Layer
\item A Convolutional Layer, 32 filters, $3\times 3$ kernel, no stride, ReLU activation function.
\item Another Convolutional Layer, same parameters. ReLU or Composite activation function. (Optional Filtering Layer 3)
\item A $2\times 2$ Pooling Layer.
\item A Densely Connected Layer with 50 Neurons and ReLU activation function.
\item A 10-classes Softmax.
\end{itemize}

The base CNN is close to the CNN used as standard ConvNet in \cite{xu_feature_2018}.

\subsection{Results}
We tested a subset of all possible candidates created in the paper. The candidates we have tested are as follow:

\subsubsection{Offset-Filtering} 
We tested only one network with an offset filtering on the filtering layer 1, using a HSAT activation function (Candidate A).

\subsubsection{Convolutional-Filtering} We tested the following Spartan Networks using the Convolutional Filtering:
\begin{itemize}
\item The standard CNN using a HSND on the layer 1, as well as Cos-HSND layers on layers 2 and 3 (Candidate B)
\item The standard CNN using a Cos-HSAT layer on layer 1 (Candidate C)
\end{itemize}

\subsubsection{Robustness}
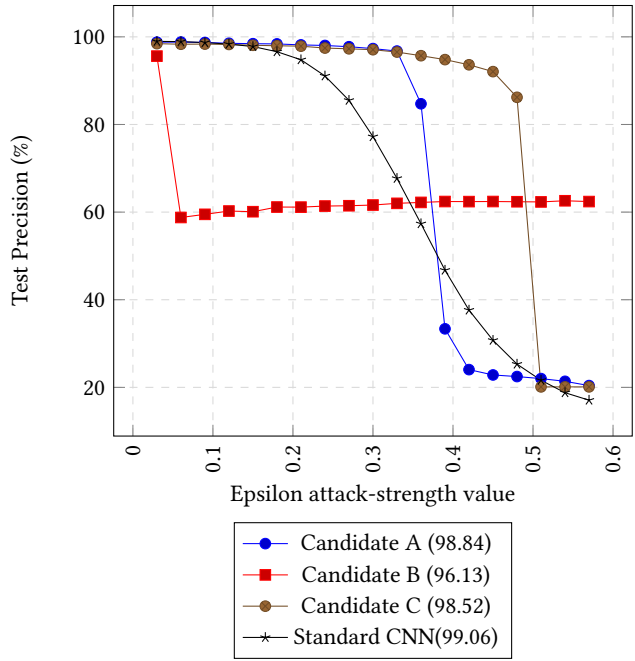
\begin{figure}[h]

  \begin{center}
    \begin{tikzpicture}
      \begin{axis}[
          width=\linewidth, 
          grid=major, 
          grid style={dashed,gray!30}, 
          xlabel=Epsilon attack-strength value , 
          ylabel=Test Precision (\%),
          legend style={at={(0.5,-0.2)},anchor=north}, 
          x tick label style={rotate=90,anchor=east} 
        ]

        \addplot 
        table[x=Epsilon,y=HSDnsRDINIT,col sep=comma] {FGSMBBDATAUG8RSLTS_stronger.csv};
        \addplot 
        table[x=Epsilon,y=SpartanCOSw4,col sep=comma] {FGSMBBDATAUG8RSLTS_stronger.csv};
        \addplot 
        table[x=Epsilon,y=SpartanHSND,col sep=comma] {FGSMBBDATAUG8RSLTS_stronger.csv};
        \legend{Candidate A (98.84), Candidate B (96.13), Candidate C (98.52), Standard CNN(99.06)}
        \addplot
        table[x=Epsilon,y=Vanilla,col sep=comma] {FGSMBBDATAUG8RSLTS_stronger.csv};
      \end{axis}
    \end{tikzpicture}
    \caption{Comparison of a Spartan Network vs its vanilla CNN counterpart, against a FGSM attack with varying epsilon-strength}
    	\label{fig:SpartanVSVanilla}
  \end{center}

\end{figure}

We attack the candidate Spartan Networks with a FGSM attack in surrogate black-box mode with various strengths. We have used the CleverHans module version 2.1.0 \cite{papernot2018cleverhans} to perform the attacks. We show our results on Fig.~\ref{fig:SpartanVSVanilla}.\\

The filter-regularization scaling factor is $s_{f}=10^{-5}$, for all of the Spartan Networks, and $\mu=0, \sigma=1$ if the Normal Distribution is used. The number in parenthesis in the legend shows the test precision (in \%) on clean samples.\\

\subsubsection{On the Training of a Network using Composite Activation Functions}
\begin{figure}[h]
\centering
\includegraphics[width=0.5\textwidth]{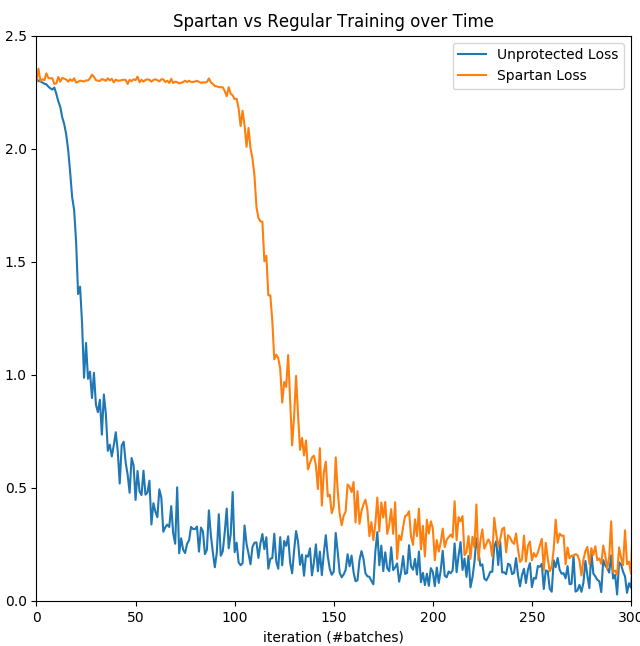}
\caption{Loss history during the training of a classical CNN and a Spartan Network. We see a latency in the loss drop of the Spartan Network's training.}
\label{fig:losstime}
\end{figure}

We report the variation of loss over training iteration of the Candidate C compared to the standard CNN using only ReLU activation functions with no filtering layers. Results are seen on Fig.~\ref{fig:losstime}.\\

As the synthetic gradient allows the Spartan Network to know the general direction of improvement, it can train even if the Heaviside step function does not yield a progressive, differentiable behaviour. After some time, the networks' weights are trained enough to go past the threshold when required, and the loss drops as sharply as a standard CNN. 

\subsubsection{Resistance to Over-capacity}
We note that in Fig.~\ref{fig:SpartanVSVanilla} Candidate C's precision begins to collapse at $\epsilon$ values close to 0.5. Note that, at this level, the noise value is high enough to create a grey picture. We see however that Candidate B manages to resist even \emph{after} the 0.5 threshold. Upon further inspection, this Network has a 0.9 cutoff threshold in its filters with close to 0 bias. This network focuses on the brightest pixels.  The attack did not supress enough bright pixels to entirely corrupt the digits, and the Spartan Networks finds the signal through the perturbation.\\

We report that out of the four given filters, the Candidate C discarded two filters by making their activation impossible. The network thus learns to ignore unnecessary capacity, by minimizing its activation penalty. This resistance is comforted by the fact that Candidate C \emph{learns} that 1-bit black and white color is enough to classify digits succesfully. This result is aligned with previous results obtained by \citet{xu_feature_2018}.

Spartan Networks thus have a reduced sensitivity to over-capacity. We argue that this resistance is trading robustness for reduced performance. This resistance assures that the hyperparameters introduced by Spartan Networks do not slow the hyperperameter tuning phase.

\section{Risk Evaluation of a Robustness-Performance tradeoff}
Spartan Networks sacrifice a bit of precision for robustness. Notions of robustness-precision tradeoffs were already mentionned in \cite{gilmer_adversarial_2018, guo_countering_2017}.\\

To give a risk analysis of a situation, consider a Spartan Network used in the context of a 4-digits check reading system, and consider an adversarial check paper that automatically gets missclassified as a 9999\$ check.  Let $\Delta_{CN\rightarrow SN}$ be the risk delta linked with going from a \emph{"ConvNet"} to a \emph{"SpartaNet"}. As the risk is the probability of occurence times the impact, we get:
$$
\Delta_{CN\rightarrow SN} = (p_{eSN}I_{eSN} + p_{tSN}I_{tSN}) - (p_{eCN}I_{eCN} + p_{tCN}I_{tCN})
$$
Where $I$ is an average impact value, $p$ a probability of an event happening, SN and CN stand for Spartan Network and Convolutional Network missclassification and t and e describe a theft scenario, and an error scenario respectively.\\

As the Network does not change the Impact, only the probability, we have:
$$
\Delta_{CN\rightarrow SN} =  (p_{eSN}-p_{eCN})I_{e}+(p_{tSN}-p_{tCN})I_{t}
$$
Each of the above probabilities is a joint event of an scenario happening, and a missclassification happening.
We consider the two events disjoint.
$$
\Delta_{CN\rightarrow SN} =  (p(e,SN)-p(e,CN))I_{e}+(p(t,SN)-p(t,CN))I_{t}
$$
We consider only abnormal check reading. $\alpha$ is the amount of malicious bank checks in the overall non-normal checks. Let $p(e,SN)-p(e,CN)=\Delta_{err}, p(t,advSN)-p(t,advCN)=\Delta_{adv}$, we get that:
$$
\Delta_{CN\rightarrow SN} =  (1-\alpha)\Delta_{err}I_{e}+\alpha \Delta_{adv}I_{t}
$$
Diminishing the risks means that $\Delta_{CN\rightarrow SN}<0$, and in this case, we obtain: 
$$
\alpha>\frac{\Delta_{err}I_{e}}{\Delta_{err}I_{e}-\Delta_{adv}I_{t}}
$$

Spartan Networks are less precise on clean samples. Thus, they are more relevant in adversarial settings. This means that there is a minimum amount of malicious checks that must be in use for the bank to have an interest in them.\\

For instance, if we hypothesize that the attack is a $\epsilon = 0.3$ FGSM attack, we see in Fig.~\ref{fig:SpartanVSVanilla} that a Spartan Network is more robust to this attack by 20\%. However, we must also consider that our Network comes with $\sim -0.5$\% precision on non-adversarial inputs. The risk analysis must take into account the fact that non-malicious checks are being misclassified more often than with a ConvNet.

\paragraph{Risk Management Implications} For example, if a non-malicious error on a 4 digits check costs on average 50\$, and a malicious error costs 8999\$ ($1000\$\rightarrow9999\$ $), Spartan Networks begin to reduce the risk when around one erroneous check on 7200 is malicious.

We thus recommend that Spartan Networks be used in conjunction with attack detection, that can use \textbf{Detection} strategies. In this setup, a classical, high-performance network would do the inferrence until an attack is detected. When such an attack happens, a Spartan Network could be used as a fallback network. In this adversarial setting, their robustness would overcome their performance issues.

\section{Conclusion}
We have presented Spartan Networks, deep neural networks that are given a data-starving layer in order to select relevant features only. The space-lowering effect allows the system to be more resistant to adversarial examples. Filtering layers reduce performance, but can, in specific threat models, be a cost-effective way to reduce an attacker's stealthiness and success rate. This robustness method is made by trying to select relevant features, thus reinforcing deep learning in its promise that feature selection can, in a long-term view, be automated.
\paragraph{Contributions}
Our contributions are the following:
\begin{itemize}
\item We introduce the concept of \emph{composite activation functions}, by separating the forward propagation and backwards propagation functions in a Deep Neural Network.
\item We introduce the idea of a \emph{self-adversarial layer}, putting an attack-agnostic layer \emph{inside} the network to starve the subsequent layers off of information.
\item We create the first \emph{Spartan Network} using the ideas above, and test its robustness to black-box attackers.
\item We evaluate the performance-robustness tradeoff of Spartan Networks in a simple threat model.
\end{itemize}

\paragraph{Future Work}

This work puts forward a lot of experiments and we encourage our fellow researchers to explore on it. \emph{Our code will be Open-Sourced to support this effort, and will be available at}:\\
\url{https://github.com/FMenet}.

We also see that the possibilities for remplacement gradients are endless:  
One could replace the gradient of a differentiable function by another to improve the update dynamic while retaining a desireable behaviour, like a gradient close to 1 to avoid exploding or vanishing gradients.\\

We expect to use more powerful, deeper and wider architectures to train on more complex datasets, but the current implementation of those networks is slow. We aim at producing a faster, more efficient way to backpropagate efficiently with our replacement gradients.\\

Image space topology allows for a smooth exploration of sample space, as a small euclidian distance often means same semantics. We leave the study of Spartan Networks on harder sample space topologies for future work.

\section*{Acknowledgments}

We would like to thank all our colleagues from the SecSI lab, and especially Ranwa Al-Mallah for useful comments and corrections.



\bibliographystyle{ACM-Reference-Format}
\bibliography{acmtog-sample-bibfile}

\end{document}